# Accuracy and Performance Comparison of Video Action Recognition Approaches

Matthew Hutchinson*, Siddharth Samsi[†], William Arcand[†], David Bestor[†], Bill Bergeron[†], Chansup Byun[†], Micheal Houle[†], Matthew Hubbell[†], Micheal Jones[†], Jeremy Kepner[†], Andrew Kirby[†], Peter Michaleas[†], Lauren Milechin[+], Julie Mullen[†], Andrew Prout[†], Antonio Rosa[†], Albert Reuther[†], Charles Yee[†], Vijay Gadepally[†]

\* MIT Department of Electrical Engineering and Computer Science, Cambridge, MA
[†] MIT Lincoln Laboratory Supercomputing Center, Lexington, MA
[+] MIT Department of Earth, Atmosphere, and Planetary Sciences, Cambridge, MA

*Abstract*—Over the past few years, there has been significant interest in video action recognition systems and models. However, direct comparison of accuracy and computational performance results remain clouded by differing training environments, hardware specifications, hyperparameters, pipelines, and inference methods. This article provides a direct comparison between fourteen "off-the-shelf" and state-of-the-art models by ensuring consistency in these training characteristics in order to provide readers with a meaningful comparison across different types of video action recognition algorithms. Accuracy of the models is evaluated using standard Top-1 and Top-5 accuracy metrics in addition to a proposed new accuracy metric. Additionally, we compare computational performance of distributed training from two to sixty-four GPUs on a state-of-the-art HPC system.

## I. INTRODUCTION

Over the past decade, the confluence of computing hardware, data availability, and algorithms have created a number of advances in machine learning and artificial intelligence [14]. For example, applications such as object recognition in images have essentially reached, and in some cases surpassed, the accuracy of human object recognition (e.g., 29 of 38 teams in the 2017 ImageNet [8] Challenge achieving error rates < 5% [15]). However, other applications such as the classification of actions in trimmed and untrimmed videos remain a challenge due to the volume and complexity of analyzing video streams. Video action recognition can be used for autonomous vehicle applications [28], reviewing security footage, organizing video databases, or summarizing video feeds.

DISTRIBUTION STATEMENT A. Approved for public release. Distribution is unlimited. This material is based upon work supported by the Under Secretary of Defense for Research and Engineering under Air Force Contract No. FA8702-15-D-0001. Any opinions, findings, conclusions or recommendations expressed in this material are those of the author(s) and do not necessarily reflect the views of the Under Secretary of Defense for Research and Engineering. © 2020 Massachusetts Institute of Technology. Delivered to the U.S. Government with Unlimited Rights, as defined in DFARS Part 252.227-7013 or 7014 (Feb 2014). Notwithstanding any copyright notice, U.S. Government rights in this work are defined by DFARS 252.227-7013 or DFARS 252.227-7014 as detailed above. Use of this work other than as specifically authorized by the U.S. Government may violate any copyrights that exist in this work.

While there were early successes using techniques such as Hidden Markov Models (HMMs) [43] or Support Vector Machines (SVMs) [38], recent innovations have largely come in the way of Deep Convolutional Neural Networks (CNNs). Deep CNNs have unique abilities to learn complex feature representations from raw data. Some algorithms trained on small video datasets with limited human action classes have demonstrated reasonable accuracy performance [22], [41], [44]; however, generalizing these results to a broader set of actions still results in high error rates [29], [44].

Unlike the field of image classification, video action recognition lacks a thorough discussion of the accuracy and computational performance metrics by which models and algorithms are compared. For example, it is difficult to compare accuracy metrics of various algorithms which are often developed and tested on heterogenous datasets or lack sufficient details regarding model architectures. Comparing a Top-5 accuracy result from a 100-class problem with a Top-5 accuracy result from a 400-class problem is impractical at best and deceptive at worst. Further, the race for higher accuracies has sidelined discussions of equally relevant aspects of model development details and training computational performance.

In this paper, we catalog a subset of state-of-the-art video action recognition models. Specifically, we

1) select a dataset most appropriate for video action recognition model comparison,
2) survey the landscape of action recognition approaches,
3) train and evaluate a set of models under similar hyperparameters, training methods, and hardware, and
4) discuss and employ accuracy and computational performance metrics.

Therefore, this paper addresses gaps in the literature and provides readers with a detailed overview of "off-the-shelf" video action recognition techniques, a novel set of action recognition accuracy metrics, a thorough side-by-side comparison of model effectiveness and efficiency, and data-driven recommendations for future use of these models.

## II. VIDEO DATASETS

Datasets for video action recognition are defined by a set of qualities: source, pre-processing, point-of-view, number of videos, length of each video, number of action classes, classes per video (single or multi-label), annotations, and purpose. A myriad of datasets have been crafted and curated to span this spectrum of qualities. This section describes the progression of these datasets and how we selected one for our experiment.

The vast majority of these datasets focus entirely on human actions due to their relevance in all aspects of everyday life. Early datasets, KTH [33], Weizmann [3], GTEA [12], GTEA GAZE [11], and GTEA GAZE+ [11], focus on daily activities.

Spurred by the growth of online video, UCF101 [35] and HMDB51 [23], with 13,000 and 7,000 videos respectively, quickly became foundational benchmarks in human action recognition. Thumos [21] and ActivityNet [19] had similar goals but did not gain the same level of popularity.

While human actions and activities continue to dominate the field of video datasets, slowly other purposes emerged. Among them, YouTube-8M [2] and MV [30] focus on broader actions and visual entities. The Something-Something [16] dataset looks at low-level action captions for intuitive physics and semantics. Moments in Time [29], a dataset with over a million 3-second videos labeled by action verb, includes both human and non-human actions.

Among the most current iterations of these datasets, only a few have the breadth (action classes) and depth (videos per class) that are comparable to ImageNet and other object recognition datasets. Kinetics-600 [22] and VLOG [13] achieve this for human actions and daily interactions. Moments in Time uniquely offers high interclass and intraclass variation. Additionally, the best performance on Moments in Time currently is a Top-1 accuracy of 38.64% and a Top-5 accuracy of 67.19% [24]. Therefore, Moments in Time is a good benchmark because of its dataset qualities and the tremendous room for improvement in models. In this paper, we use Moments in Time training and validation data.

Moments in Time was first used in the CVPR'18 action recognition challenge which tasked teams to develop state-of-the-art methods for achieving high Top-1 and Top-5 accuracies. However, the optionally submitted reports by the top-performing teams lack key details of implementation and training with almost no mention of hyperparameter selection, validation methods, hardware, or computational performance. By rigorously describing the experimental setup, this paper provides grounds of model comparison that cannot be easily derived from the original Moments in Time paper nor the 2018 Moments in Time Challenge results.

## III. CLASSIFICATION TECHNIQUES

Action recognition models can be broadly arranged into 2D and 3D approaches referring to their 2-dimensional or 3-dimensional convolutional kernels, respectively. Figure 1 displays four of the most common architectures. This section describes these approaches and provides insights obtained from those employed in the Moments in Time challenges.

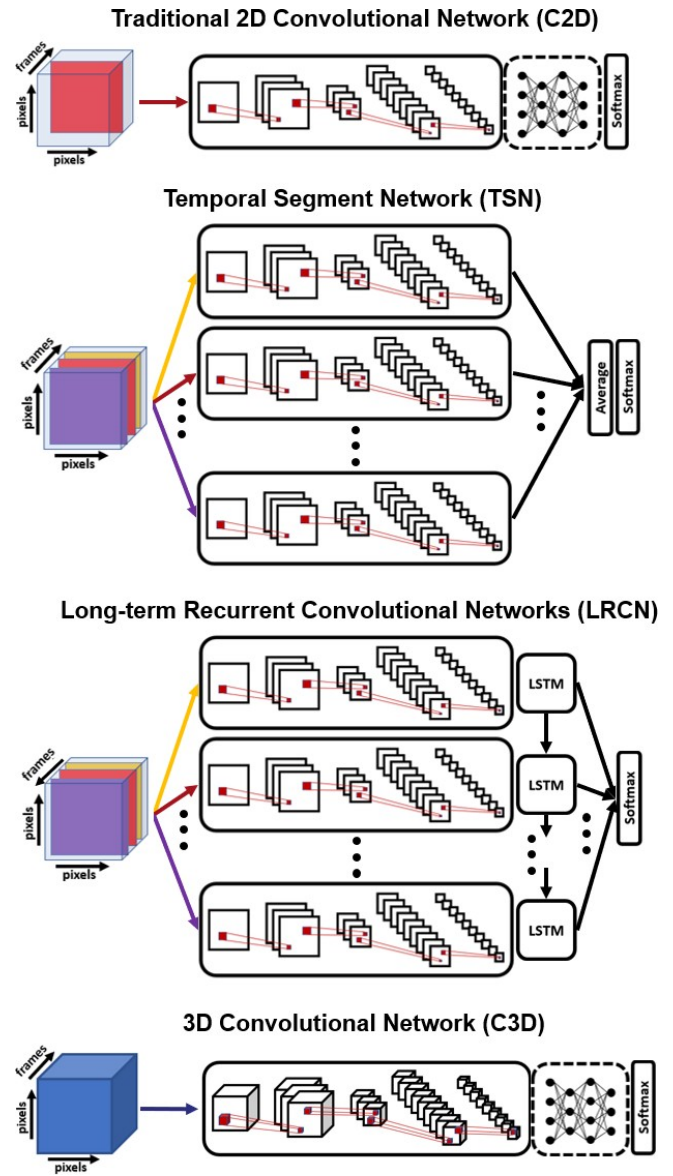

Fig. 1. Common video action recognition approaches. Note that these are simplified where icons for 2D-ConvNets, 3D-ConvNets, dense classification networks, LSTM modules, and averaging/softmax layers are used only as visual interpretation. Their actual design can vary significantly. Similarly, the majority of approaches have 2-stream variants for RGB+Optical Flow.

### A. 2D Approaches

2D approaches include traditional 2D Convolutional Neural Networks (C2D), Temporal Segment Networks (TSN) [41], Long-term Recurrent Convolutional Neural Networks (LRCN) [10] sometimes referred to as CNN+LSTMs, and Temporal Shift Modules (TSM) [27]. C2D models are derived directly from the image recognition field. With C2D, a frame is extracted from the video and used as input to a 2D-ConvNet. After several convolution and pooling layers, the logits are fed into one or more fully-connected layers which produce a softmax output prediction over the dataset classes. With TSN, a video is segmented along its temporal (frames) dimension and one frame is extracted from each segment for input to 2D-ConvNets that share weights. The predictions from each segment are then averaged before the softmax output layer. Variants and additions to TSN include Temporal Relations Networks (TRN) [44] which performs multi-scale relationing. LRCN also segments a video, extracts a frame from each segment and feeds those frames into 2D-ConvNets. However, the ConvNet outputs are used as inputs to a Long-Short Term Memory (LSTM) network prior to softmax predictions.

TABLE I
MOMENTS IN TIME CHALLENGE 2018 TOP-PERFORMING MODELS

| Model Type | Model Backbone | Validation Accuracy Top-1 % | Top-5 % | report |
|---|---|---|---|---|
| *Video Models* | | | | |
| C2D | SENet152 | **33.7** | 61.3 | [25] |
| | SEResNeXt | 30.0 | 60.2 | [25] |
| | Xception | 31.8 | 59.2 | [25] |
| | ResNet50 | 28.3 | 53.2 | [25] |
| TSN | ResNet152 | **33.0** | n/a | [40] |
| | DPN107 | 31.1 | n/a | [40] |
| | ResNet50 | 27.4 | 53.2 | [25] |
| TRN | SENet154 | **31.9** | 58.8 | [6] |
| | Inception-v3 | 29.7 | 55.7 | [6] |
| | InceptionResNet-v2 | 29.3 | 55.6 | [6] |
| I3D | ResNet50 | **34.2** | 61.4 | [25] |
| | ResNet101-NL | 33.7 | n/a | [40] |
| | Inception-v3 | 27.6 | 53.9 | [6] |
| C3D | InceptionResNet-v2 | **35.1** | 63.3 | [24] |
| | ResNet101 | 33.6 | 61.2 | [24] |
| *Audio Models* | | | | |
| C2D | VGGish | **17.1** | n/a | [40] |
| | SENet50 | 16.8 | n/a | [40] |
| | M34-res | 14.8 | 27.4 | [24] |
| | ResNet34 | 13.8 | 23.6 | [24] |
| | EnvNet+ResNet | 13.2 | 25.9 | [24] |
| | NetVLAD | 9.0 | 19.5 | [6] |
| | SoundNet | 7.6 | 18.0 | [26] |

*B. 3D Approaches*

3D-Convolutional Neural Networks (C3D) were designed as the 3D analogy to 2D-ConvNets [37]. However, because of the long-term dependencies of actions, C3D models often have had less success on action recognition than their 2D counterparts on object recognition. To attempt to bridge the gap between 2D and 3D models, Inflated 3D (I3D) models were created by "inflating" pretrained 2D kernels into 3D kernels [5]. This allows I3D models to benefit from pretraining on 2D image datasets like ImageNet. Some believe that, while still in their early days, 3D approaches will be able to retrace the successful history of their 2D siblings [17]. Both C3D and I3D use either the entire video or a selected portion (e.g. 16, 32, or 64 frames) as an input to a 3D-ConvNet. Similar to the C2D approach, the 3D-ConvNet's output is then fed into a classification network before outputting softmax predictions.

*C. Application to Moments in Time*

Table I shows the best performing RGB-mode single-model variants from these 2D and 3D approach categories in the 2018 Moments in Time Challenge. Most teams tackled the multi-modal problem (RGB, optical flow, audio) with a late-fusion/ensemble of several models. Teams placing within the top 10 ensembled on average 9 individual models together.

While model architectures such as C3D, I3D and TRN may intuitively be expected to provide significant advantages, from our survey of the 2018 challenge results, these architectures barely outperform, and in some cases underperform, C2D and TSN approaches. This combined with the "off-the-shelf" nature of many of these models and pretrained weights makes them prime for the bulk of this comparison study.

The Multi-Moments in Time Challenge (ICCV'19) tasked teams to detect multiple event labels from videos. Of the optional reports that teams submitted, few deviated from 2D approaches likely because they drew similar conclusions as those described above and encountered difficulties of working on more complex models. Both challenges also demonstrated that the incorporation of audio into late-fusion ensembles had little to no additive benefits over only using the video stream. It is hypothesized that the majority of the action classes (e.g. *running, jumping, catching*) in Moments in Time are more easily expressed visually than in audio.

## IV. METRICS

Video datasets are often orders of magnitude larger than image datasets. A single video in Moments in Time contains 90 frames and one or two audio tracks. This increases the storage and computational requirements for training. Therefore, while often overlooked in the literature, including both accuracy and computational performance in model assessment is instrumental in comparing video action recognition approaches. This section describes the metrics we use for these comparisons.

*A. Accuracy Metrics*

Top-1 and Top-5 percentage accuracy metrics are canonically used for comparison in action recognition. We argue that plotting what we call a pseudo-Receiver Operator Characteristic (p-ROC) curve as shown in Figure 2 (or any simple transformation of it) allows for more intuitive metrics of accuracy comparison because it shows all values of $k \in \{0, ..., |C|\}$ where $C$ is the set of action classes (i.e. $C=\{applauding, baking, crashing, descending\}$ in the case of Moments in Time). In this p-ROC, the Top-$k$ accuracy is plotted against $k$ analogous to plotting the true positive rate against the false positive rate for our classifier.

Even though it has been noted that the ROC area under the curve ($AUC$) and the maximum Youden index ($J_{max}$), the curve height above the chance line, provide desirable properties as a classification metric [4], [20], the practice has not become standard. These benefits easily transfer to our p-ROC curve and allow a user to quickly and more intuitively select a model with accuracy characteristics that they desire. Equations 1 and 2 show how to calculate p-ROC $AUC$ and $J_{max}$ where $acc(k)$ refers to a function computing Top-$k$ accuracy for a given $k$.

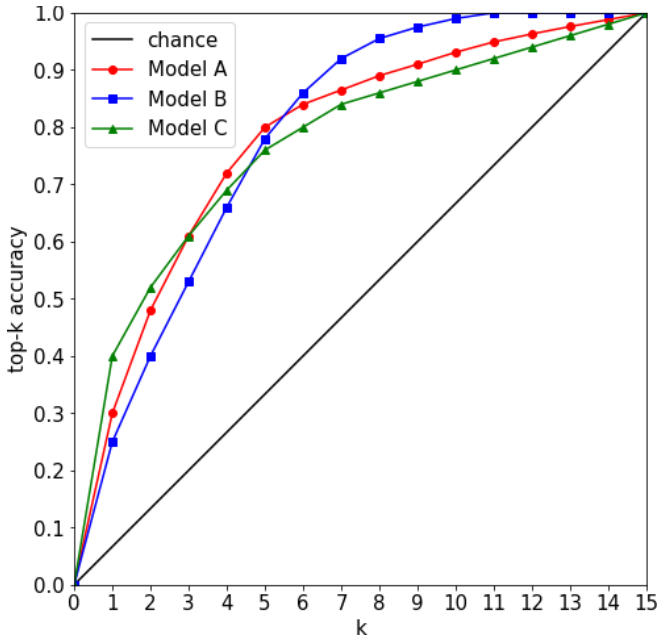

Fig. 2. Example p-ROC curve for a 15-class problem. Three model Top-k values are plotted (as well as a random chance line that represents an uninformed guesser.

## TABLE II
## MODEL DEPTH AND COMPLEXITY IN COMPARISON

| Model Type | Model Backbone | Layers | Trainable Parameters |
|---|---|---|---|
| C2D | VGG19 | 19 | 20,198,291 |
| | MobileNet (M) | 28 | 3,554,451 |
| | Inception-v3 (Iv3) | 48 | 22,462,963 |
| | ResNet50 (R50) | 50 | 24,229,203 |
| | MobileNet-v2 (Mv2) | 53 | 2,658,131 |
| | Xception (X) | 71 | 21,501,563 |
| | Inception-ResNet-v2 (IRv2) | 164 | 54,797,235 |
| | DenseNet169 (D169) | 169 | 13,048,915 |
| | DenseNet201 (D201) | 201 | 18,744,147 |
| LRCN | n/a (16f) | 38 | 9,788,915 |
| C3D | n/a (16f) | 18 | 148,590,675 |
| | n/a (32f) | 18 | 456,872,019 |
| I3D | Inception-v1 (Iv1) (16f) | 27 | 12,279,984 |
| | Inception-v1 (Iv1) (64f) | 27 | 12,279,984 |

$$AUC = \sum_{k=0}^{|C|-1} \frac{acc(k+1) - acc(k)}{2} \quad (1)$$

$$J_{max} = \max_{k \in \{0,1,...,|C|\}} acc(k) - \frac{k}{|C|} \quad (2)$$

We claim that there are two key benefits to using p-ROC curves as the basis for reporting action recognition model accuracy results. First, metrics derived from p-ROC, such as $AUC$, provide a more holistic view of accuracy across all Top-k values rather than at a specific k. Real-world applications of action recognition will undoubtedly require different accuracy requirements, and p-ROC provides the model user a wider picture of the model's capabilities. If only Top-1 and Top-5 accuracies were reported for the example shown in Figure 2, one might naively conclude that Model *C* or Model *A* is the "best". However, Model *B* actually has the highest p-ROC *AUC* of 11.82 compared with Model *A*'s 11.72 and Model *C*'s 11.56. Second, p-ROC metrics can be dataset agnostic. $J_{max}$ is by definition independent of the number of classes, and p-ROC *AUC* can be easily normalized as shown in Equation 3. Therefore, model accuracies can be compared across datasets with different numbers of classes.

$$AUC_{norm} = \widehat{AUC} = \frac{AUC}{|C|} \quad (3)$$

### B. Computational Performance Metrics

Because of the data-scale challenges of dealing with video, computational performance of training is essential in assessing models. In this paper, we measure computational performance directly by training time and training time per epoch. Attention was also paid to how varying the compute resources affects training (i.e. yields speedup curves).

## V. EXPERIMENTAL DESIGN

This section describes the setup for comparing fourteen TensorFlow [1] implementations of action recognition models. Those include nine C2D models (VGG19, MobileNet, Inception-v3, ResNet50, MobileNet-v2, Xception, Inception-ResNet-v2, DenseNet169, and DenseNet201 [1]), one LRCN model [18], two C3D models [36], and two I3D models [9]. Table II describes the depth and complexity of these models.

### A. Software

Python 3.6.5 scripts trained and validated these "off-the-shelf" models in a distributed fashion using Horovod 0.18.2 [34] and OpenMPI 4.0. Key package versions used were NumPy 1.16.5 [31], H5py 2.9.0 [7], SciPy 1.3.2 [39], TensorFlow 1.14.0 [1], and FFmpeg 3.3.7.

### B. Hardware

Models were trained on 1, 2, 4, 8, 16, and 32 nodes. Each node consists of 2x20 Intel Xeon G6-6248 CPU cores with two NVIDIA Volta V100 GPUs (PCIe connection) and has 384 GB RAM and 3.8 TB local disk space. The infrasturture used is described in detail in [32].

### C. Pre-Processing

The Moments in Time pre-processed 30 frames per second (fps) videos resized to 224x224 frames are used as an input. We parse videos by extracting the frames to NumPy arrays and vertically stacking the frames to create 3D video "cubes". This was completed for training and validation sets that were defined by the Moments in Time creators.

### D. Training and Validation

C2D and I3D models were initialized with ImageNet pre-trained weights while C3D and LRCN models were initialized with random weights. On each pass through the dataset during training, C2D inputs were randomly sampled frames from each video. LRCN, C3D (f16), and I3D (f16) had inputs of 16 evenly spaced frames. C3D (32f) and I3D (64f) randomly sampled 32 and 64 continuous frames, respectively.

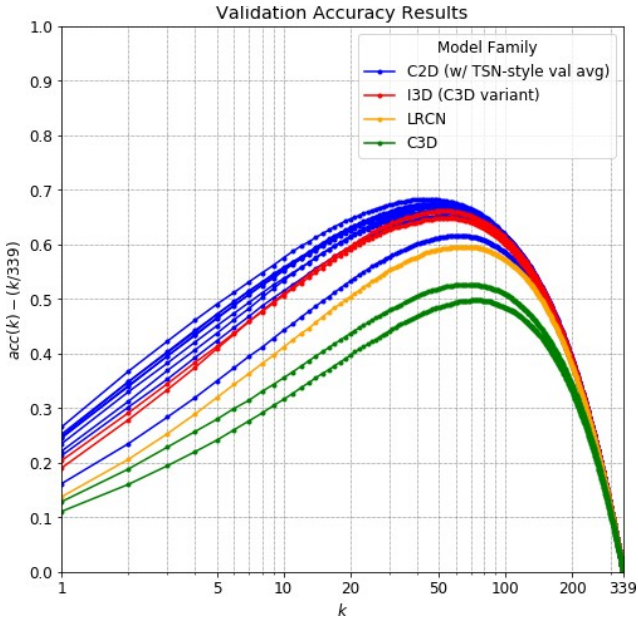

Fig. 3. p-ROC curve log-scaled with k/339 subtracted out for each value of $k$ to more easily show the peak $J$-statistic

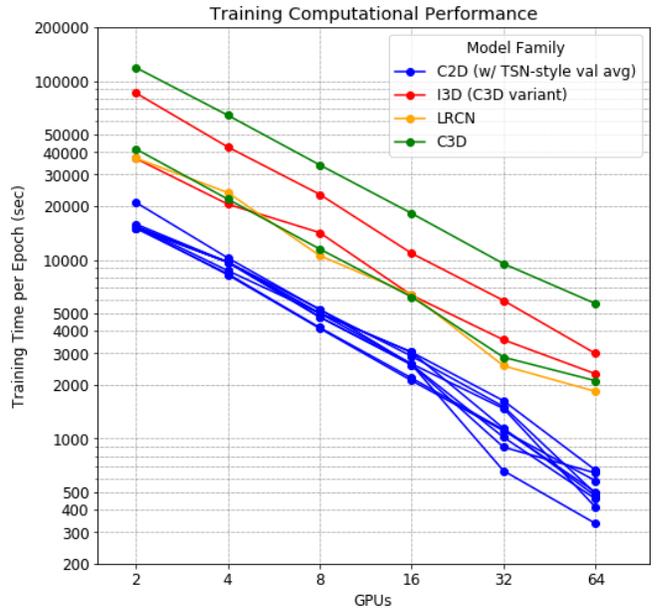

Fig. 4. Training Time per Epoch across training configurations of 1, 2, 4, 8, 16, and 32 nodes where each node as 2 Volta V100 GPUs.

For proper comparison given the computational resources limitations of training 14 models, other hyperparameters were held consistent across different training sessions. A Horovod-wrapped distributed ADADELTA [42] optimizer and categorical cross-entropy loss metric updated network weights. Five warmup epochs slowly raised the learning rate to 1.0 which was subsequently decayed at 20, 35, and 50 epochs. Each model was trained for 65 epochs. Early stopping was not possible due to the distributed training environment. For validation, LRCN, C3D, and I3D model inference was performing the same as training. C2D model inference was performed in a TSN-style averaging across 6 evenly spaced frames for the 90 frame (30 fps) video.

## VI. RESULTS AND DISCUSSION

Accuracy results and computational performance results are summarized in Figures 3 and 4, respectively, as well as in Table III. Figure 5 describes the relation between accuracy and computation performance observed in this experiment.

### A. Accuracy Results

By our p-ROC accuracy metrics, the top three performing models were all C2Ds: Inception-ResNet-v2, Xception, and DenseNet169 with $AUC$ validation results of 0.919, 0.919, and 0.917, respectively. These models are noted for their greater depth (i.e. more layers). We also note that Inception-ResNet-v2 was the highest performer using the traditional Top-1 and Top-5 accuracy metrics. Among 2D approaches, the LRCN yielded the worst results across all accuracy metrics reported likely due to its shallower model backbone depth.

Of the 3D approaches, the 16 frame I3D-Inception-v1 model yielded the best validation accuracy results. However, none of the 3D models surpassed the six best performing C2D models tested when comparing p-ROC $AUC$ results.

It is interesting to note that the maximum Youden $J$-statistics ($J_{max}$) are occurring between $k=41$ and $k=74$ for these models which indicates the very arbitrary nature of Top-5 accuracy as a metric. While not as pronounced in this comparison study, other experiments we have conducted demonstrate that not only is it possible for Top-5 accuracy to be deceptive, but it is actually common for it to be uncorrelated with increased or decreased $AUC$ and $J_{max}$.

### B. Computational Performance Results

As expected, essentially all models approximately halved their training times when trained on twice as many GPUs. Minor node differences and network lag on the system became more apparent in the larger (16 and 32 node) training runs as evidenced by increased variation on the right of Figure 4.

When trained on 64 GPUs, the three quickest trained models were ResNet50, Inception-ResNet-v2, and MobileNet-v2 with training times of 335.4, 413.5, 460.7 seconds per epoch, respectively. Those correspond to total training times of 6.06, 7.47, and 8.32 hours. C2D models trained between 3x and 17x faster than the 3D models in this experiment. LRCN, while having fewer layers and trainable parameters than other 2D models tested, had noticeably higher training times due to the 16 frame input size. Clearly, there are computational costs to be paid for models with larger inputs.

TABLE III
ACCURACY AND COMPUTATIONAL PERFORMANCE RESULTS

| Model Type | Backbone | Val Acc (%) Top-1 | Top-5 | p-ROC metrics $\widehat{AUC}$ | $J_{max}$ | Training Time per Epoch (s) on $g$ Volta V100 GPUs $g=2$ | $g=4$ | $g=8$ | $g=16$ | $g=32$ | $g=64$ |
|---|---|---|---|---|---|---|---|---|---|---|---|
| | random chance | 0.29 | 1.47 | 0.5 | 0.0 | | | | | | |
| C2D | VGG19 | 16.45 | 36.41 | 0.891 | 0.616 ($k=59$) | 15256.1 | **8235.5** | **4135.0** | **2109.6** | 1106.4 | 580.8 |
| | M | 21.61 | 43.79 | 0.908 | 0.652 ($k=51$) | **15103.4** | 9693.4 | 4783.7 | 2585.2 | 895.6 | 642.5 |
| | Iv3 | 24.87 | 48.20 | 0.915 | 0.673 ($k=48$) | 20920.4 | 10238.6 | 5253.3 | 2890.2 | 1509.8 | 493.0 |
| | R50 | 23.77 | 46.54 | 0.911 | 0.661 ($k=47$) | 15261.6 | 9694.5 | 5298.6 | 2594.9 | **660.8** | **335.4** |
| | Mv2 | 22.42 | 45.16 | 0.915 | 0.670 ($k=50$) | **15103.4** | 9687.1 | 4783.7 | 2592.3 | 1017.8 | 460.7 |
| | X | 24.84 | 47.58 | 0.919 | 0.683 ($k=41$) | 14997.3 | 8342.6 | 4177.8 | 2179.3 | 1120.7 | 502.1 |
| | IRv2 | **26.83** | **50.50** | **0.919** | **0.683** ($k=41$) | 15831.6 | 9694.6 | 5019.1 | 2619.3 | 1473.0 | 413.5 |
| | D169 | 25.13 | 48.63 | 0.917 | 0.674 ($k=45$) | 15399.2 | 9690.7 | 5019.1 | 3041.9 | 1627.4 | 666.6 |
| | D201 | 25.52 | 48.62 | 0.915 | 0.672 ($k=46$) | 15399.4 | 8687.1 | 5019.1 | 3041.9 | 1141.9 | 478.9 |
| LRCN | n/a (16f) | 14.04 | 33.40 | 0.883 | 0.596 ($k=63$) | 37009.4 | 23740.9 | 10553.6 | 6388.7 | 2553.8 | 1835.8 |
| C3D | n/a (16f) | 13.15 | 29.41 | 0.842 | 0.499 ($k=74$) | 41622.7 | 21823.9 | 11485.8 | 6195.0 | 2851.1 | 2107.2 |
| | n/a (32f) | 11.36 | 25.58 | 0.824 | 0.499 ($k=74$) | 118738.2 | 64250.2 | 33911.3 | 18177.1 | 9505.5 | 5688 |
| I3D | Iv1 (16f) | 19.33 | 42.36 | 0.911 | 0.661 ($k=52$) | 36838.8 | 20456.2 | 14182.7 | 6331.1 | 3567.3 | 2303 |
| | Iv1 (64f) | 20.69 | 42.74 | 0.904 | 0.649 ($k=57$) | 85565.7 | 42697.6 | 23209.8 | 10864.0 | 5916.8 | 2991.8 |

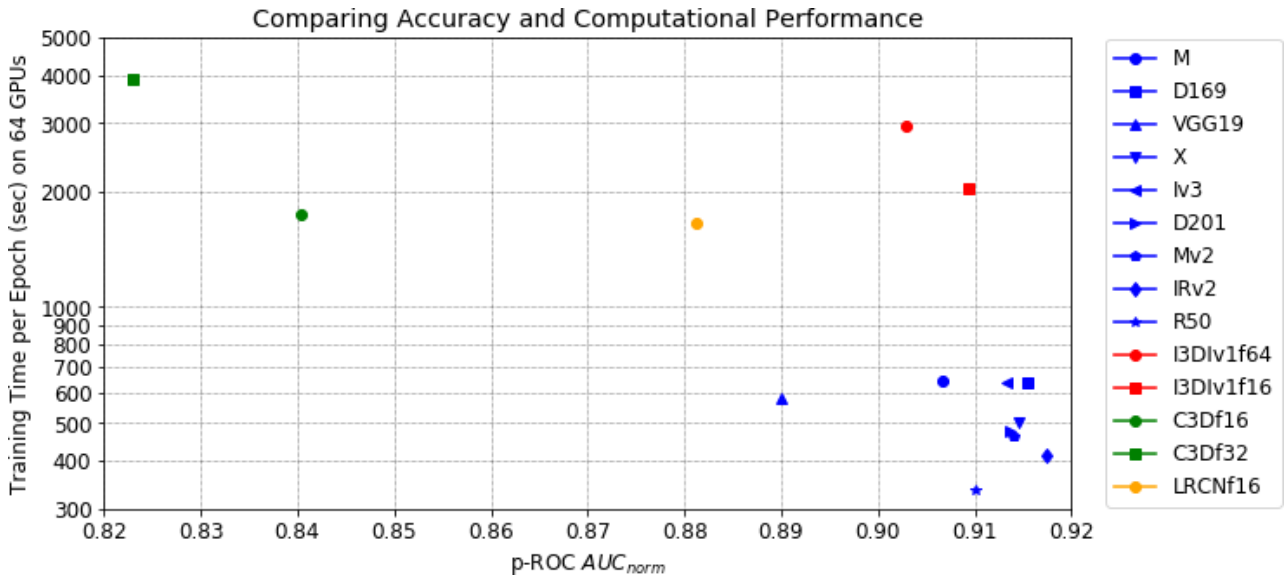

Fig. 5. A plot of training time per epoch (measured in seconds) against p-ROC $\widehat{AUC} = AUC_{norm}$ to show accuracy and computational performance trade-offs. The best performing models are found in the lower-right corner.

## VII. CONCLUSION

When holistically looking at both accuracy and computational performance, as shown in Figure 5, C2D models significantly outperform the LRCN, C3D, and I3D models tested in this study. Of the C2D models, those found in the bottom right of the plot—Inception-ResNet-v2, ResNet50, DenseNet201, and MobileNetv2—are the top performers.

These results concur with the 2018 Moments in Time Challenge observations that 2D approaches can yield results comparable to their more complex 3D counterparts, and model depth, rather than input feature scale, is the critical component to an architecture's ability to extract a video's semantic action information. For approaching a new action recognition problem, we therefore recommend starting with C2D models with TSN-style inference. We also recommend utilizing the p-ROC dataset agnostic accuracy metrics ($AUC$ and $J_{max}$) presented in Section IV of this paper.

This study has also benchmarked the computational costs of training these models. As expected, 3D models clearly incur significant computational costs alluded to, but rarely clarified, in the literature. The results we have presented should offer a training time baseline for these models across a variety of computational resource scales on state-of-the-art hardware.

Future comparison work should expand the list of compared models into other 2D and 3D backbones as well as benchmark training times on alternative high performance computing (HPC) infrastructures. Access to more computational resources would allow for hyperparameter tuning. Future action recognition research should focus on model depth and not dismiss the application of simpler C2D models.

REFERENCES

[1] M. Abadi, A. Agarwal, P. Barham, E. Brevdo, Z. Chen, C. Citro, G. S. Corrado, A. Davis, J. Dean, M. Devin, S. Ghemawat, I. Goodfellow, A. Harp, G. Irving, M. Isard, Y. Jia, R. Jozefowicz, L. Kaiser, M. Kudlur, J. Levenberg, D. Mané, R. Monga, S. Moore, D. Murray, C. Olah, M. Schuster, J. Shlens, B. Steiner, I. Sutskever, K. Talwar, P. Tucker, V. Vanhoucke, V. Vasudevan, F. Viégas, O. Vinyals, P. Warden, M. Wattenberg, M. Wicke, Y. Yu, and X. Zheng. TensorFlow: Large-scale machine learning on heterogeneous systems, 2015. Software available from tensorflow.org.

[2] S. Abu-El-Haija, N. Kothari, J. Lee, P. Natsev, G. Toderici, B. Varadarajan, and S. Vijayanarasimhan. Youtube-8m: A large-scale video classification benchmark. *arXiv preprint arXiv:1609.08675*, 2016.

[3] M. Blank, L. Gorelick, E. Shectman, M. Irani, and R. Basri. Actions as space-time shapes. *2005 IEEE International Conference on Computer Vision (ICCV)*, 1:1395–1402, 2005.

[4] A. P. Bradley. The use of the area under the ROC curve in the evaluation of machine learning algorithms. *Pattern Recognition*, 30(7):1145–1159, July 1997.

[5] J. Carreira and A. Zisserman. Quo vadis, action recognition? A new model and the kinetics dataset. *2017 IEEE Conference on Computer Vision and Pattern Recognition (CVPR)*, 2017.

[6] C. Chen, X. Wei, X. Zhao, and Y. Liu. Alibaba-venus at activitynet challenge 2018 – task C trimmed event recognition (moments in time). *http://moments.csail.mit.edu/challenge2018/Alibaba_Venus.pdf*, 2018.

[7] A. Collette. *Python and HDF5*. O'Reilly Media, November 2013.

[8] J. Deng, W. Dong, R. Socher, L. Li, K. Li, and L. Fei-Fei. Imagenet: A large-scale hierarchical image database. *2009 IEEE Conference on Computer Vision and Pattern Recognition (CVPR)*, pages 248–255, 2009.

[9] dlpbc. Keras implementation of inflated 3d from quo vardis paper + weights. MIT. *https://github.com/dlpbc/keras-kinetics-i3d*, 2017.

[10] J. Donahue, L. A. Hendricks, S. Guadarrama, M. Rohrbach, S. Venugopalan, K. Saenko, and T. Darrell. Long-term recurrent convolutional networks for visual recognition and description. *2015 IEEE Conference on Computer Vision and Pattern Recognition (CVPR)*, 2015.

[11] A. Fathi, Y. Li, and J. M. Rehg. Learning to recognize daily actions using Gaze. *2012 European Conference on Computer Vision (ECCV)*, pages 314–327, 2012.

[12] A. Fathi, X. Ren, and J. M. Rehg. Learning to recognize objects in egocentric activities. *2011 IEEE Conference on Computer Vision and Pattern Recognition (CVPR)*, pages 3281–3288, 2011.

[13] D. F. Fouhey, W. Kuo, A. Efros, and J. Malik. From lifestyle vlogs to everyday interactions. *2018 IEEE Conference on Computer Vision and Pattern Recognition (CVPR)*, pages 4991–5000, 2018.

[14] V. Gadepally, J. Goodwin, J. Kepner, A. Reuther, H. Reynolds, S. Samsi, J. Su, and D. Martinez. AI enabling technologies: a survey. *arXiv preprint axXiv:1905.03592*, 2019.

[15] D. Gershgorn. The Quartz guide to artificial intelligence: What is it, why is it important, and should we be afraid? *https://qz.com/1046350/the-quartz-guide-to-artificial-intelligence-what-is-it-why-is-it-important-and-should-we-be-afraid/*, September 2017.

[16] R. Goyal, S. E. Kahou, V. Michalski, J. Materzynska, S. Westphal, H. Kim, V. Haenel, I. Fründ, P. Yianilos, M. Mueller-Freitag, F. Hoppe, C. Thurau, I. Bax, and R. Memisevic. The "Something Something" video database for learning and evaluating visual common sense. *2017 IEEE International Conference on Computer Vision (ICCV)*, pages 5843–5851, 2017.

[17] K. Hara, H. Kataoka, and Y. Satoh. Can spatiotemporal 3D CNNs retrace the history of 2D CNNs and ImageNet? *2018 IEEE Conference on Computer Vision and Pattern Recognition (CVPR)*, pages 6546–6555, 2018.

[18] Matt Harvey. Five video classification methods implemented in Keras and TensorFlow. Medium. *https://blog.coast.ai/five-video-classification-methods-implemented-in-keras-and-tensorflow-99cad29cc0b5*, March 2017.

[19] F. C. Heilbron, V. Escorcia, B. Ghanem, and J. C. Niebeles. ActivityNet: A large-scale video benchmark for human activity understanding. *2015 IEEE Conference on Computer Vision and Pattern Recognition (CVPR)*, pages 961–970, 2015.

[20] J. Huang and C. Ling. Using AUC and accuracy in evaluating learning algorithms. *IEEE Transactions on Knowledge and Data Engineering*, 17:299–310, March 2005.

[21] H. Idrees, A. R. Zamir, Y. Jiang, A. Gorban, I. Laptev, R. Sukthankar, and M. Shah. The THUMOS challenge on action recognition for videos "in the wild". *Computer Vision and Image Understanding (CVIU)*, 155, April 2016.

[22] W. Kay, J. Carreira, K. Simonyan, B. Zhang, C. Hillier, S. Vijayanarasimhan, F. Viola, T. Green, T. Back, P. Natsev, M. Suleyman, and A. Zisserman. The Kinetics human action video dataset. *arXiv preprint arXiv:1705.06950*, 2017.

[23] H. Kuehne, H. Jhuang, E. Garrote, T. Poggio, and T. Serre. HMDB: A large video database for human motion recognition. *2011 International Conference on Computer Vision (ICCV)*, pages 2256–2563, 2011.

[24] C. Li, Z. Hou, J. Chen, Y. Bu, J. Zhou, Q. Zhong, D. Xie, and S. Pu. Team DEEP-HRI Moments in Time challenge 2018 technical report. *http://moments.csail.mit.edu/challenge2018/DEEP_HRI.pdf*, 2018.

[25] Y. Li, Z. Xu, Q. Wu, Y. Cao, S. Zhang, L. Song, J. Jiang, C. Gan, G. Yu, and C. Zhang. Team Megvii submission to Moments in Time challenge 2018. *http://moments.csail.mit.edu/challenge2018/Megvii.pdf*, 2018.

[26] Z. Li and L. Yao. Team UNSW-Data Science submission to the Moments in Time challenge 2018. *http://moments.csail.mit.edu/challenge2018/UNSW_Data_Science.pdf*, 2018.

[27] J. Lin, C. Gan, and S. Han. Temporal shift module for efficient video understanding. *2019 IEEE/CVF International Conference on Computer Vision (ICCV)*, pages 7082–7092, 2019.

[28] O. Moll, A. Zalewski, S. Pillai, S. Madden, M. Stonebraker, and V. Gadepally. Exploring big volume sensor data with vroom. *Proceedings of the VLDB Endowment*, 10(12):1973–1976, 2017.

[29] M. Monfort, B. Zhou, S. A. Bargal, A. Andonian, T. Yan, K. Ramakrishnan, L. M. Brown, Q. Fan, D. Gutfreund, C. Vondrick, and A. Oliva. Moments in Time dataset: one million videos for event understanding. *IEEE Transactions on Pattern Analysis and Machine Intelligence (TPAMI)*, 42(2):502–508, February 2020.

[30] P. X. Nguyen, G. Rogez, C. C. Fowlkes, and D. Ramanan. The open world of micro-videos. *arXiv preprint arXiv:1603.09439*, 2016.

[31] T. Oliphant. *Guide to NumPy*, 2nd edition. CreateSpace Independent Publishing Platform:. September 2015.

[32] A. Reuther, J. Kepner, C. Byun, S. Samsi, W. Arcand, D. Bestor, B. Bergeron, V. Gadepally, M. Houle, M. Hubbell, M. Jones, A. Klein, L. Milechin, J. Mullen, A. Prout, A. Rosa, C. Yee, and P. Michaleas. Interactive supercomputing on 40,000 cores for machine learning and data analysis. *2018 IEEE High Performance Extreme Computing Conference (HPEC)*, pages 1–6, 2018.

[33] C. Schuldt, I. Laptev, and B. Caputo. Recognizing human actions: A local SVM approach. *2004 International Conference on Pattern Recognition (ICPR)*, 3:32–36, 2004.

[34] A. Sergeev and M. Del Balso. Horovod: fast and easy distributed deep learning in TensorFlow. *arXiv preprint arXiv:1802.05799*, 2018.

[35] K. Soomro, A. R. Zamir, and M. Shah. UCF101: A dataset of 101 human actions classes from videos in the wild. *arXiv preprint arXiv:1212.0402*, 2012.

[36] axon-research. C3D for Keras + TensorFlow. TASER International, Inc. *https://github.com/axon-research/c3d-keras/blob/master/LICENSE.md*, 2017.

[37] D. Tran, L. Bourdev, R. Fergus, L. Torresani, and M. Paluri. Learning spatiotemporal features with 3D convolutional networks. *2015 IEEE International Conference on Computer Vision (ICCV)*, pages 4489–4497, 2015.

[38] J. Uijlings, I. C. Duta, E. Sangineto, and N. Sebe. Video classification with densely extracted hog/hof/mbh features: An evaluation of the accuracy/computational efficiency tradeoff. *International Journal of Multimedia Information Retrieval (IJMIR)*, 4(1):33–44, 2015.

[39] P. Virtanen, R. Gommers, T. E. Oliphant, M. Haberland, T. Reddy, D. Cournapeau, E. Burovski, P. Peterson, W. Weckesser, J. Bright, S. J. van der Walt, M. Brett, J. Wilson, K. Jarrod Millman, N. Mayorov, A. R. J. Nelson, E. Jones, R. Kern, E. Larson, C.J. Carey, İ. Polat, Y. Feng, E. W. Moore, J. Vand erPlas, D. Laxalde, J. Perktold, R. Cimrman, I. Henriksen, E. A. Quintero, C. R. Harris, A. M. Archibald, A. H. Ribeiro, F. Pedregosa, P. van Mulbregt, and SciPy 1. 0 Contributors. SciPy 1.0: Fundamental Algorithms for Scientific Computing in Python. *Nature Methods*, 2020.

[40] Z. Xiaoteng, B. Yixin, Z. Feiyun, H. Kai, W. Yicheng, Z. Liang, H. Qinzhu, L. Yining, S. Jie, and P. Yao. Team Qiniu submission to ActivityNet Challenge 2018. *http://moments.csail.mit.edu/challenge2018/Qiniu.pdf*, 2018.

[41] Y. Xiong, Z. Wang, Y. Qiao, D. Lin, X. Tang, and L. Van Gool. Temporal segment networks: Towards good practices for deep action recognition. *2016 European Conference on Computer Vision (ECCV)*, pages 20–36, 2016.

[42] M. D. Zeiler. ADADELTA: an adaptive learning rate method. *arXiv preprint arXiv:1212.5701*, 2012.

[43] H. Zhang. Video action recognition based on hidden markov model combined with particle swarm. *International Journal on Computer Science and Information Systems (IADIS)*, 7(2):1–17, 2012.

[44] B. Zhou, A. Andonian, A. Oliva, and A. Torralba. Temporal relation reasoning in videos. *2018 European Conference on Computer Vision (ECCV)*, pages 803–818, 2018.